\documentclass[10pt,twocolumn,letterpaper]{article}

\usepackage{iccv}
\usepackage{times}
\usepackage{epsfig}
\usepackage{graphicx}
\usepackage{amsmath}
\usepackage{amssymb}
\usepackage{tabularx}
\usepackage{multirow}

\usepackage[breaklinks=true,bookmarks=false]{hyperref}

\iccvfinalcopy 

\def\iccvPaperID{****} 

\ificcvfinal\fi

\begin{document}

\title{Efficient, Self-Supervised Human Pose Estimation with Inductive Prior Tuning}

\author{Nobline Yoo\\
Princeton University\\
{\tt\small nobliney@alumni.princeton.edu}
\and
Olga Russakovsky\\
Princeton University\\
{\tt\small olgarus@princeton.edu}
}

\maketitle
\ificcvfinal\thispagestyle{empty}\fi

\begin{abstract}
    The goal of 2D human pose estimation (HPE) is to localize anatomical landmarks, given an image of a person in a pose. SOTA techniques make use of thousands of labeled figures (finetuning transformers or training deep CNNs), acquired using labor-intensive crowdsourcing. On the other hand, self-supervised methods re-frame the HPE task as a reconstruction problem, enabling them to leverage the vast amount of unlabeled visual data, though at the present cost of accuracy. In this work, we explore ways to improve self-supervised HPE. We (1) analyze the relationship between reconstruction quality and pose estimation accuracy, (2) develop a model pipeline that outperforms the baseline which inspired our work, using less than one-third the amount of training data, and (3) offer a new metric suitable for self-supervised settings that measures the consistency of predicted body part length proportions. We show that a combination of well-engineered reconstruction losses and inductive priors can help coordinate pose learning alongside reconstruction in a self-supervised paradigm.

\end{abstract}

\section{Introduction}\label{ch:intro}

Applications of human pose estimation span a wide range, including predicting pedestrian behavior and trajectory on roads with autonomous vehicles \cite{bauer_2023_weakly, kress_2019_human, wang_leverage_2019, zanfir_2023_hum3dil, Zheng_2022_multimodal}. SOTA works \cite{geng_human_2023, xu_vitpose_2022} have explored larger models or tackled specific failure modes (\eg occlusion). Broadly speaking, some of the latest HPE models address one or more of the following five questions. (1) Can we speed up prediction to enable real-time pose estimation \cite{garau_capsulepose_2023, li_human_2022}? (2) Can we create lightweight models with smaller memory footprints \cite{li_human_2022, osokin_real-time_2018, santavas_attention_2021, xu_can_2023, zhang_simple_2020}? (3) Pose estimation models lack robustness under situation $X$. How can we address this \cite{geng_human_2023, yang_camerapose_2023, zhang_neuromorphic_2023}? (4) Can we use transformers \cite{einfalt_uplift_2023, qian_hstformer_2023, xu_vitpose_2022, zhao_dpit_2022}? (5) Vision-based pose estimation is unreliable. Can we use other more robust modes of data \cite{geng_densepose_2022, lee_hupr_2023}?

\begin{figure}[t]
    \centering
            \includegraphics[width=\linewidth]{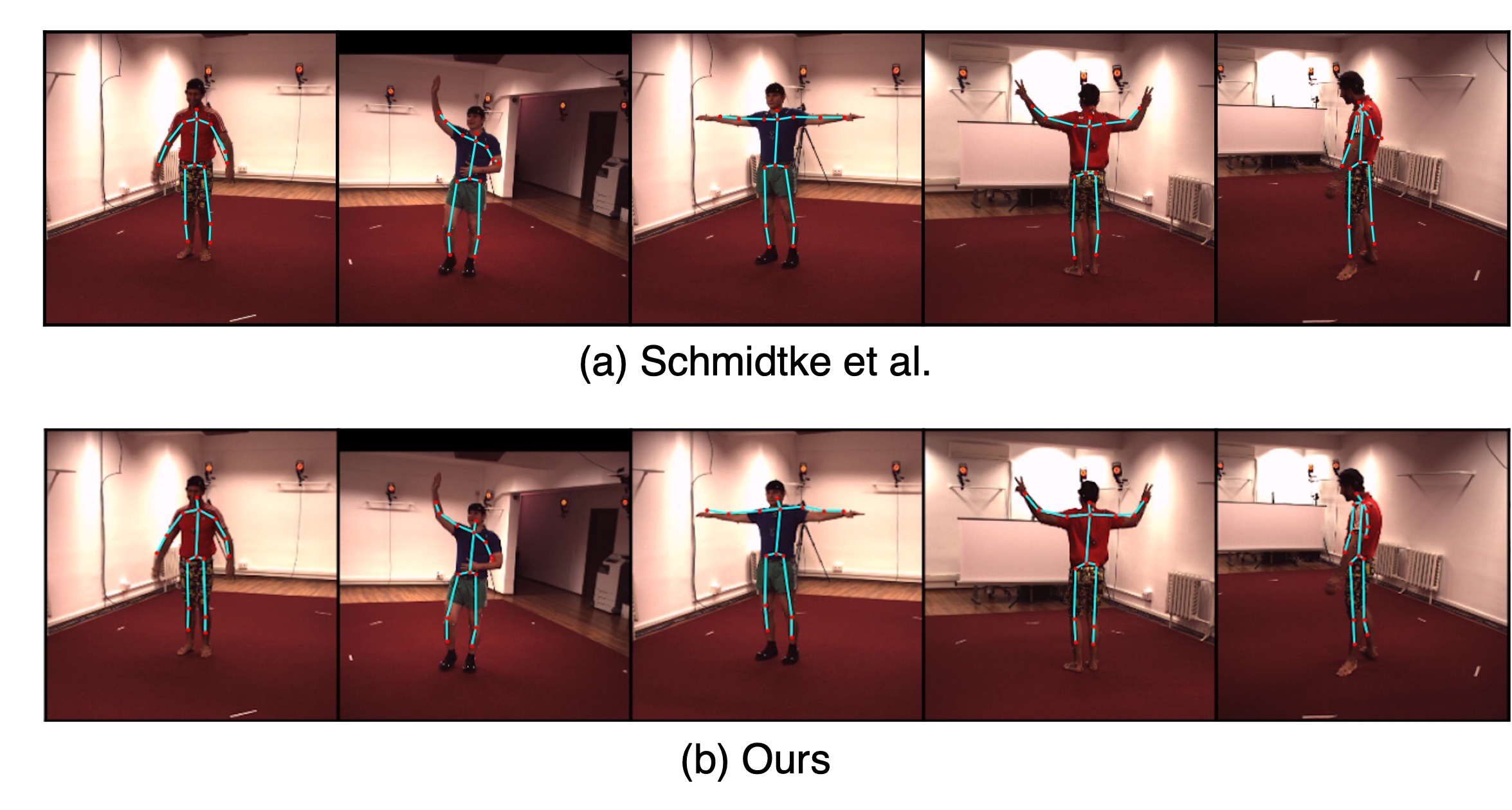}
\caption{\textbf{(a) Baseline predictions} of~\cite{schmidtke_unsupervised_2021}. \textbf{(b) Ours} (with MSE loss, $\boldsymbol{T_{new}}$ template, coarse-to-fine learning). Our predictions more closely follow the outline of the subjects.}
\label{fig:1}
\end{figure}

An important caveat is that many of these methods require labor-intensive pose labeling, which limits scalability in training data. To illustrate, COCO's training and validation set alone contain 1.7 million labeled keypoints \cite{lin2014microsoft}. MPII Human Pose contains more than 600,000 \cite{andriluka_2d_2014}.

To make use of the vast amount of unlabeled data, we look to self-supervised models, which frame classification and regression as reconstruction problems \cite{cao_-bed_2022, li_geometry-driven_2020, schmidtke_unsupervised_2021, wan_self-supervised_2019}, where given some parts of the input space (source), the model reconstructs other parts of the input space (target). These methods are engineered such that classification or regression are necessary to reason about the signals absent in the source but present in the target to be recovered. Hence, rather than optimizing directly for the task, self-supervision learns indirectly by optimizing for reconstruction, yielding representations that surpass the generalizability of those learned via supervised learning \cite{tendle_study_2021}. Self-supervised learning holds much potential for autonomous driving, facilitating greater scalability in training data, improving robustness, and enabling lifelong-learning \cite{berscheid2020self, deng2020self}. SOTA methods in this space typically rely on multi-view geometry \cite{bouazizi2021self, gholami2022self, Kocabas_2019_self, Wandt_2021_CVPR}, unpaired pose data \cite{Kundu_2020_self, sosa2023self}, or synthetic datasets with pose labels \cite{Kundu_2022_uncertainty}.

\begin{figure*}[t]
\centering
\includegraphics[width=\linewidth]{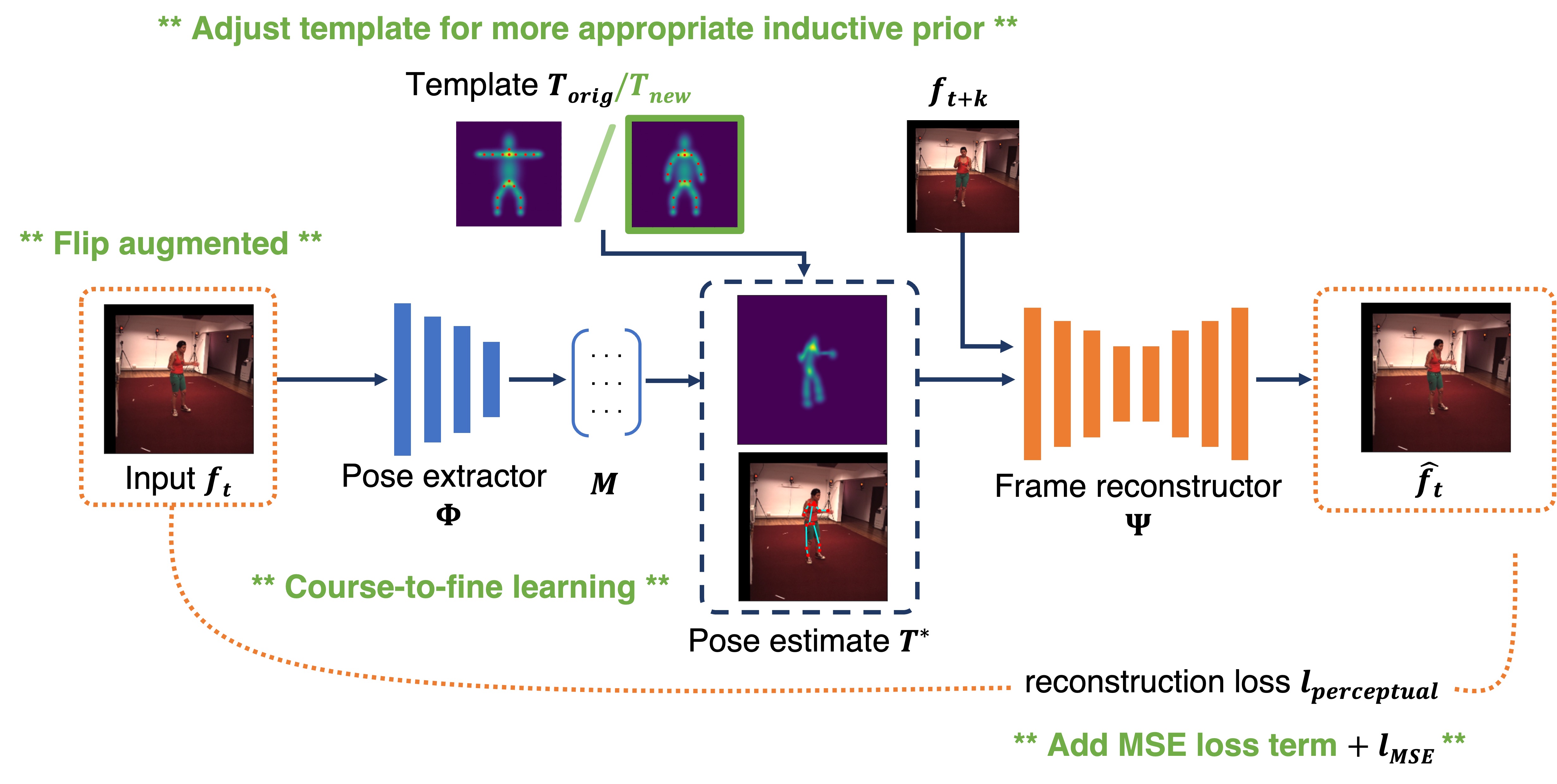}
\caption{\textbf{Baseline model architecture.} Our adjustments to the pipeline are in \textbf{green}.}
\label{fig:2}
\end{figure*}

Recently, Schmidtke et al.~\cite{schmidtke_unsupervised_2021} use a template-based, self-supervised approach to estimate pose in the Human3.6M dataset \cite{ionescu_human36m_2014}, without relying on multi-view geometry, unpaired pose data, or synthetic datasets with pose labels. By guiding the model with templates of Gaussians, each representing a body part, the model is able to render feasible pose estimates. However, without labels, there is no guarantee that predictions will reflect ground truth \cite{shu_feature-metric_2020}. 

In this work, we analyze the relationship between reconstruction optimization and pose estimate accuracy in the absence of ground truth, using a carefully engineered combination of reconstruction loss and inductive prior (in the form of a new template). Using these insights,  we develop an efficient model pipeline that exceeds the baseline \cite{schmidtke_unsupervised_2021} performance, using a training set that is less than one-third the size of the original, along with a new formulation of matrix transformation that implements coarse-to-fine learning and data augmentation. Further, we propose a metric of consistency for body part lengths that is suitable for self-supervised settings, where ground truth is not available. Our code is available at \url{https://github.com/princetonvisualai/hpe-inductive-prior-tuning}.

\section{Approach}

\subsection{Baseline}
\label{sec:baseline}

We begin by describing the baseline model \cite{schmidtke_unsupervised_2021} which we build off of. It consists of pose extractor $\Phi$ and frame reconstructor $\Psi$ (encoder-decoder), and a template $T_{orig}$ of 18 Gaussians (one per body part), which form the inductive prior for viable human shapes (Figure \ref{fig:2}). Network $\Phi$ is made up of 14 fully convolutional layers, followed by two fully-connected layers. Reconstructor $\Psi$ consists of an encoder and decoder, each of which contain seven fully-convolutional layers.

Network $\Phi$ takes input frame $f_t$, which contains the pose to estimate, and outputs transformation matrices $M = \{M_{1 \leq i \leq 18}\}$. $MT_{orig}$ generates pose estimate $T^*$. Reconstructor $\Psi$ takes $T^*$ along with frame $f_{t+k}$ (a frame with the same background and subject as $f_t$) to reconstruct $f_t$ by jointly reasoning over the subject style and background information present in $f_{t+k}$ and pose information from $T^*$.

The model is trained using three objectives: (1) an anchor-point loss to keep adjacent body parts together, (2) a boundary loss to keep the predicted keypoints within the physical 256$\times$256 frame, and (3) a perceptual reconstruction loss with a VGG backbone pretrained on ImageNet to reflect human judgements of similarity \cite{ledig_photo-realistic_2017}. $l_{anchor}$ and $l_{boundary}$ are as defined in \cite{schmidtke_unsupervised_2021}.
\begin{gather}
    l_{recon} = ||VGG(\hat{f_t}) - VGG(f_t)||_1^1 
    \label{equ:lrecon_orig}\\
    l_{tot} = l_{recon} + \lambda_1 l_{anchor} + \lambda_2 l_{boundary}     \label{equ:ltot_orig}
\end{gather}

In baseline predictions, the chest, shoulder, and knee keypoints are relatively collapsed, while the general form only loosely follows the subject's outline (Figure \ref{fig:1}a). Limb lengths are not always consistent; the second and third frame in Figure \ref{fig:1}a depict the same subject from approximately the same camera angle; however, the length of the forearm is significantly reduced in the former, presenting an issue of inconsistency.

\subsection{Change 1: MSE reconstruction loss}
\label{sec:reconstructionloss}

Since self-supervised HPE uses reconstruction optimization as a proxy for pose learning, we consider how different formulations of reconstruction loss might improve pose estimates. While earlier works measure perceptual against pixel-wise loss \cite{ghodrati_mr_2019, johnson_perceptual_2016, snell_learning_2017}, recent works in reconstruction use pixel-wise $L_1$ or $L_2$ in conjunction with perceptual loss \cite{sharma2022neural, wang_multiscale_2003, zhang_deblurring_2020}. So, we experiment with adding an MSE term to $l_{recon}$ in Equation~\ref{equ:lrecon_orig} to improve reconstruction, with $l_{MSE}$  computed across all pixels and the RGB color channels. The reconstruction loss is now
\begin{equation}
    l_{recon} = l_{MSE} + ||VGG(\hat{f_t}) - VGG(f_t)||_1^1
\end{equation}

\subsection{Change 2: New template $T_{new}$}
\label{sec:newtemplate}

\begin{figure}[t]
    \centering
        \includegraphics[width=\linewidth]{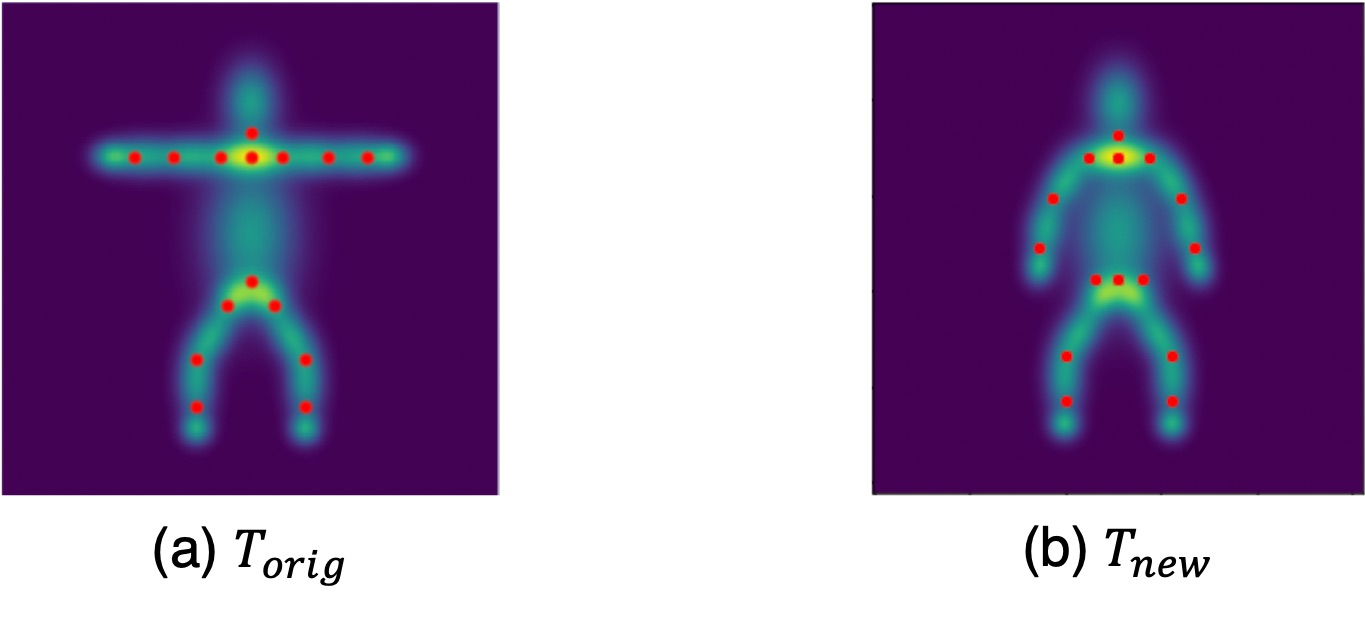}
\caption{\textbf{(a) $\boldsymbol{T_{orig}}$.} \textbf{(b) $\boldsymbol{T_{new}}$.} The new template better reflects our data distribution.}
\label{fig:16}
\end{figure}

We engineer a template $T_{new}$ 
shown in Figure \ref{fig:16}b that better reflects the natural distribution of poses in our dataset, particularly the arms-down pose, providing a more appropriate inductive prior.

\subsection{Change 3: Coarse-to-fine learning}
\label{sec:coarsetofinelearning}

Previous works have used multiple networks for coarse-to-fine learning to reconstruct finer details \cite{richardson_learning_2017, zeng_df2net_2019}. 
Instead of using multiple networks, we modify the last layer of network $\Phi$; since matrix $M$ directly influences the model's representation capability, we expand $M$ to hold 20 matrices ($M = \{M_{1 \leq i \leq 20}\}$) to estimate pose in two steps. By expanding the dimensions of matrix $M$ and selectively mapping $M_i$ (in a one-to-one or one-to-many manner), we enable coarse-to-fine learning of pose details without designing a separate network for fine reconstruction. Note that we only apply this two-step procedure to the arms, forearms, and hands, since this is where the $+$MSE, $T_{new}$ model has the most room for improvement (Figure \ref{fig:32}).

Concretely, previously, transformation matrices $M = \{M_{1\leq i \leq 18}\}$. Instead, we expand $M$ to $\{M_{1\leq i \leq 14 + 6}\}$. In step 1, the first 14 matrices $M_{1 \leq i \leq 14}$ transform $T_{new}$ to coarse estimate $T^{*'}$. In particular, $M_{10}$ and $M_{11}$ are applied to the whole left and right arm, respectively. Effectively, we treat each whole arm as a course unit. In step 2, the last six matrices $M_{15 \leq i \leq 20}$ dictate finer transformations of the individual components of the left and right arms (upper arm, forearm, hands) to get from $T^{*'}$ to final estimate $T^*$. Specifically, we apply $M_{1\leq i \leq 20}$ as in Table~\ref{table:twostepwarp1}.

\begin{table}[t]
\begin{center}
\begin{tabular}{|c|c|}
\hline
$\boldsymbol{M_{i}}$ & \textbf{Apply $\boldsymbol{M_{i}}$ to...}\\
\hline
$M_{1}$ & Core\\
\hline
$M_{2}, M_{3}$ & Left/Right hip\\
\hline
$M_{4}, M_{5}$ & Left/Right thigh\\
\hline
$M_{6}, M_{7}$ & Left/Right shin\\
\hline
$M_{8}, M_{9}$ & Left/Right shoulder\\
\hline
$M_{10}$ & \textit{Left upper arm, forearm, hand}\\
\hline
$M_{11}$ & \textit{Right upper arm, forearm, hand}\\
\hline
$M_{12}, M_{13}$ & Left/Right foot\\
\hline
$M_{14}$ & Head\\
\hline
\hline
$M_{15}, M_{18}$ & Left/Right upper arm\\
\hline
$M_{16}, M_{19}$ & Left/Right forearm\\
\hline
$M_{17}, M_{20}$ & Left/Right hand\\
\hline
\end{tabular}
\end{center}
\caption{\textbf{Coarse-to-fine learning.} Steps 1 and 2 in coarse-to-fine pose estimation.}
\label{table:twostepwarp1}
\end{table}

\subsection{Change 4: Flip dataset augmentation}
\label{sec:datasetaugmentation}
We adopt a simple dataset augmentation approach to further improve model training. Concretely, as is standard in image classification, we augment the dataset by flipping the input images across the longitudinal axis to overcome potential discrepancies in distribution between the two sides. 

\subsection{Change 5: Constraining for consistency}
\label{sec:newconsistency}
Finally, this brings us to our last contribution of a new metric that helps better constrain the model predictions. Resuming our discussion on issues of inconsistent limb lengths in baseline predictions (Section~\ref{sec:baseline}), we note that in self-supervised HPE, it becomes important to explicitly code for those things which are automatically coded for in their fully-supervised counterparts via ground truth labels; consistency is one such example. To this day, definitions of ``consistency" in HPE have largely been kept to the 3D setting, where it is defined as consistency of 3D representation across camera views for the same subject and pose and is tackled by creating new, reprojection losses \cite{gholami2022self, li_geometry-driven_2020, wan_self-supervised_2019, Wandt_2021_CVPR}. Others have directly addressed inconsistency in limb lengths by learning limb length priors in the 3D supervised setting \cite{ma2021context}.

We propose a metric for consistency in body part length proportions across frames that can be used in self-supervised settings with no ground truth labels (in the 2D setting, but it can be extended to the 3D setting). We define body part length proportion (\texttt{BPLP}) as the proportion of predicted limb length to torso length (Equation \ref{equ:BPLP}). The central motivation is that a single subject maintains a level of consistent \texttt{BPLP} across different poses. We formulate \texttt{BPLP} consistency (\texttt{BPLP-C}) as the reciprocal of the standard deviation in \texttt{BPLP}s per body part (Equation \ref{equ:BPLPC}). A higher \texttt{BPLP-C} is indicative of more consistent predictions.
\begin{gather}
    \texttt{BPLP}(\text{limb } i)=\frac{\text{predicted length(limb } i\text{)}}{\text{predicted length(torso)}} \label{equ:BPLP}\\
    \texttt{BPLP-C} = \frac{1}
    {\frac{1}{n}\sum_{1\leq i \leq n} \sigma_{\texttt{BPLP},i}} \label{equ:BPLPC}\\
    \sigma_{\texttt{BPLP},i}: \text{std. dev. of } \texttt{BPLP}\text{(limb } i \text{) over text examples}\nonumber\\
    n\text{: number of limbs}\nonumber
\end{gather}

We note how ground truth labels are not needed to calculate \texttt{BPLP} or \texttt{BPLP-C}, making them suitable for evaluating models in the self-supervised setting. To improve \texttt{BPLP-C}, we work on constraining $M$, since $M$ influences the space of pose estimates. We see this experiment as a proof-of-concept to improve \emph{consistency}, rather than as a method to improve specific performance metrics, like PDJ or $L_2$ error. We approach the problem as follows. For simplicity, we assume that \texttt{BPLP}s are consistent across all subjects. We recognize this does not hold in the real-world; but for simplicity, we work with this assumption. In concluding, we briefly discuss how we can do away with this assumption in future work by computing subject-specific \texttt{BPLP}s.

The major theoretical constraint we introduce is to apply scaling together on all limbs, since the size of individual limbs does not change drastically and disproportionately across poses, relative to other limbs, for a single subject. Hence, we have a new transformation matrix $M_i$ that is derived as follows:
\begin{gather}
    M_i=RLS\\
    M_i=\begin{bmatrix}
            cos(\theta) & sin(\theta) & 0\\
            -sin(\theta) & cos(\theta) & 0\\
            0 & 0 & 1
            \end{bmatrix}
        \begin{bmatrix}
            1 & 0 & \mu\\
            0 & 1 & \delta\\
            0 & 0 & 1
            \end{bmatrix}
         \begin{bmatrix}
            \phi & 0 & 0\\
            0 & \beta & 0\\
            0 & 0 & 1
            \end{bmatrix} \nonumber\\
    \qquad\qquad\qquad\quad R \qquad\qquad\qquad\;\; L \qquad\qquad\;\; S\qquad\;\nonumber\\
    T^* = MT 
\end{gather}

In this constrained setting, $T$ should be a subject-specific template, since $M$ is not free to scale individual limbs, but in this paper we assume a generic template. $R$ (rotation) and $L$ (localization) are for rotating and repositioning individual limbs, respectively. $S$ (scaling) is for scaling all limbs together.

$\theta, \mu, \delta$ are limb-specific parameters, while $\phi$ and $\beta$ are frame-specific parameters, in that for a given frame with a subject in some pose, there will be 3 parameters for each limb and 2 frame-specific parameters that apply to all the limbs. Hence, in the original baseline setting, there would be $(18\times3)+2 = 56$ salient, transformation parameters, whereas in the coarse-to-fine setting, there would be $(20\times3)+2 = 62$ such parameters.

\section{Experiments}

\subsection{Setup}

Similar to the baseline~\cite{schmidtke_unsupervised_2021}, we train our models on Human3.6M subjects 1, 5, 6, 7, and 8, evaluate on subjects 9 and 11, and group frames based on subject and background. We downsample frames to 256$\times$256. To encourage training efficiency, we keep the training set relatively small in all experiments---around 180K pairs of frames $(f_t, f_{t+k})$. Our model is trained over 50 epochs on 1 NVIDIA A100/V100 with a learning rate of 0.001, batch size of 48, $\lambda_1 = 0.5$, and $\lambda_2 = 1$. Training typically takes 34 hours; we evaluate our model on 130K images.

\subsection{Evaluation}
We use two overall evaluation metrics on a set $J$ of 15 keypoints: Percentage of Detected Joints (PDJ) (Equation~\ref{equ:pdj}) and $L_2$ error normalized for frame size (Equation~\ref{equ:norm_l2_error}).
\begin{gather}
    \text{PDJ@0.05} = \frac{1}{|J|} \sum_{j \in J} f(j, \hat{j}) \label{equ:pdj}\\
    f(j, \hat{j}) = 1 \text{ if dist}(j,\hat{j}) \leq .05 \times \small\text{diagonal person length} \\
    \text{Per-joint accuracy } = \frac{1}{\small\text{\# test instances}} \sum_{\text{test instances}} f(j, \hat{j})\\
    L_2 \text{ error} = \frac{1}{|J|} \sum_{j \in J} \frac{\text{dist}(j,\hat{j})}{\text{image size}} \label{equ:norm_l2_error}
\end{gather}

The model detects joint $j$ if estimate $\hat{j}$ is within 0.05 of the diagonal length of the person bounding box. A higher PDJ and lower $L_2$ error are characteristics of a more accurate model. As with the baseline, we do not predict pose orientation. We hope to address this in future work. In the meantime, we use a frame-centric lens to describe joint-handedness. Before we explore our modifications to the pipeline, we analyze the baseline in the next section.

\subsection{Baseline results}
First, please note that there are reproducibility issues in the baseline's codebase (which we confirmed with the authors). Going forward, we distinguish between the ``baseline'' (obtained from the codebase), upon which we implement our proposed changes, and the published checkpoint, against which we ultimately compare our best model.

\begin{figure}[t]
    \centering
    \includegraphics[width=\linewidth]{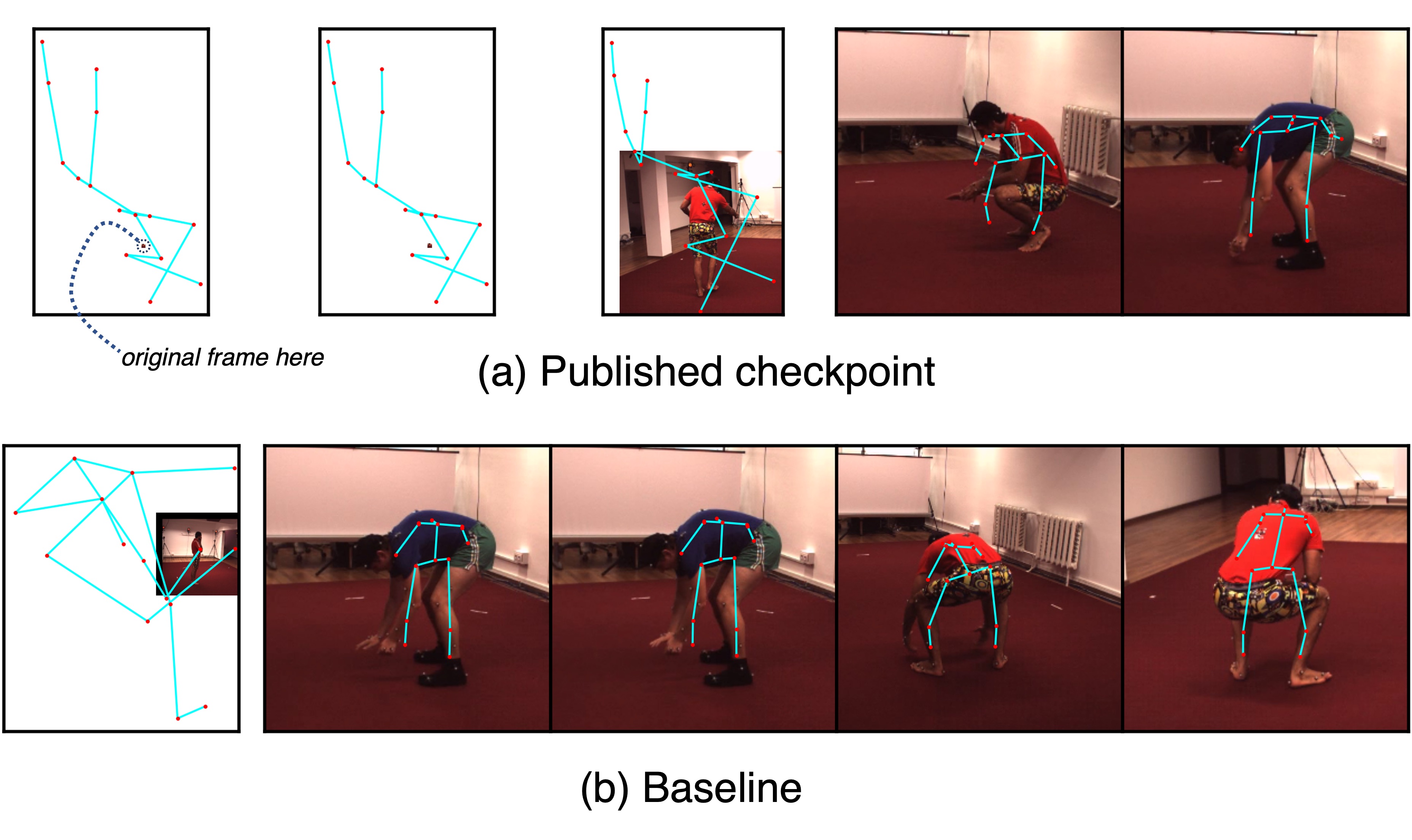}
\caption{\textbf{The bottom 0.5\% of worst predictions from the published checkpoint and baseline.} \textbf{(a)} In the first two, the predicted keypoints are so out-of-bounds; the original frame is barely visible. \textbf{(b)} The baseline yields less severe outliers.}
\label{fig:11}
\end{figure}

\begin{figure}[h]
    \centering
        \includegraphics[width=\linewidth]{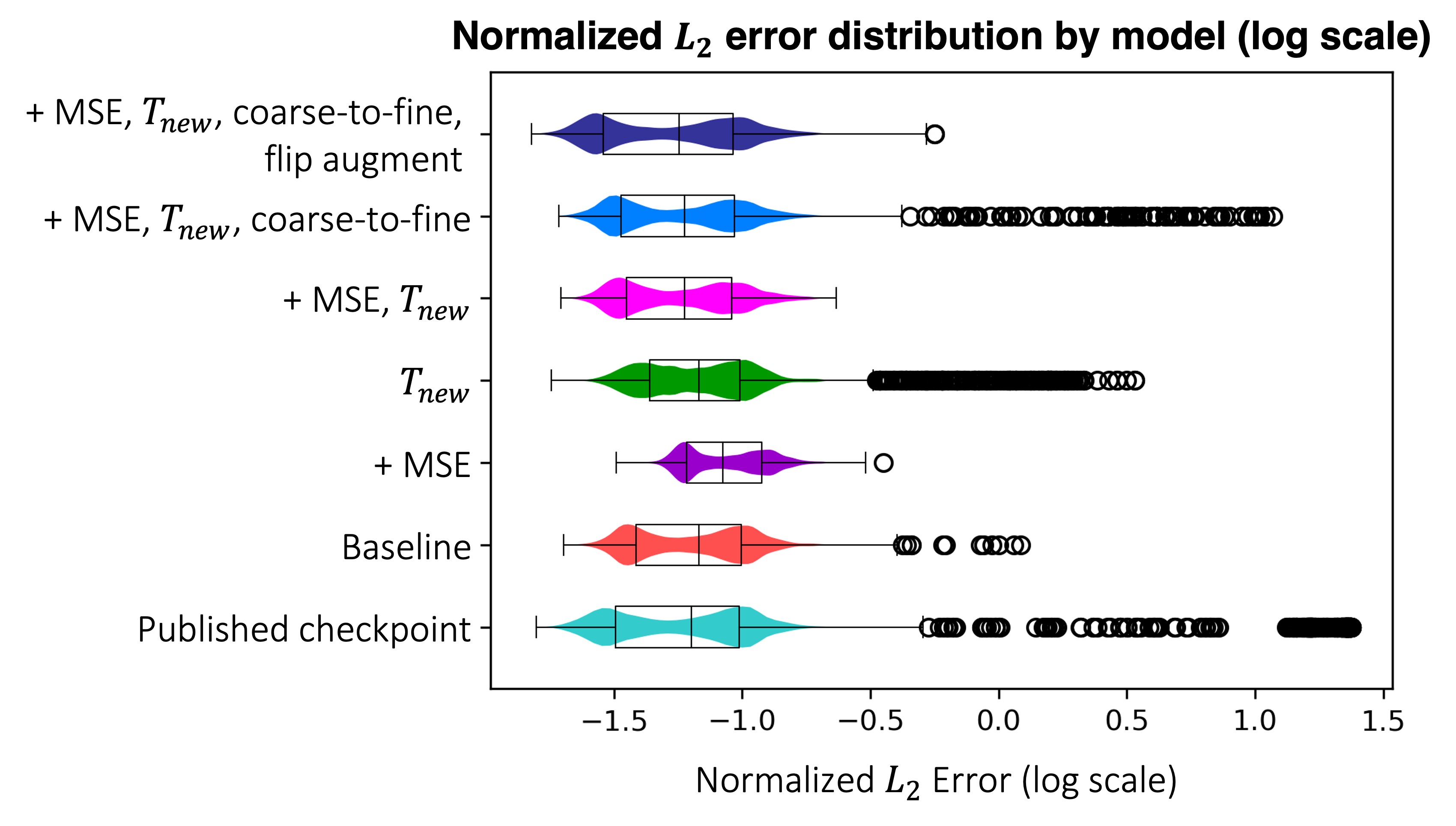}
        \caption{\textbf{Error by model.} We compute the normalized $L_2$ error (on a log scale) per frame and plot its distribution by model.}
        \label{fig:8}
\end{figure}

The baseline yields a PDJ@0.05 of 38.5 and normalized $L_2$ error of 7.2 (Table \ref{table:metrics}). Surprisingly, the baseline has a better $L_2$ error than the published checkpoint, despite a slightly worse PDJ. Figure \ref{fig:11} shows the published checkpoint yields more outliers, with keypoints that overshoot the image frame boundary. Quantitatively, Figure \ref{fig:8}, which shows the distribution of $L_2$ by model, confirms the same.

\begin{table}[t]
\begin{center}
\begin{tabular}{|m{4.5cm}|c|c|}
\hline
\multicolumn{1}{|c|}{\textbf{Model}} & \textbf{PDJ} & $\boldsymbol{L_2}$ \textbf{Error} \\
\hline
Published checkpoint & 40.8 & 11.0 \\
\hline
Baseline & 38.5 & 7.2 \\ 
\hline
$+$MSE & 26.6 & 9.2\\
\hline
$+T_{new}$ & 33.1 & 7.5\\
\hline
$+$MSE, $T_{new}$ & 37.2 & 6.7\\
\hline
$+$MSE, $T_{new}$, flip augment & 34.3 & 6.9\\
\hline
$+$MSE, $T_{new}$, coarse-to-fine & 39.0 & 7.0\\
\hline
$+$MSE, $T_{new}$, coarse-to-fine, flip augment & \textbf{42.6} & \textbf{6.4} \\
\hline
$+$MSE, $T_{new}$, coarse-to-fine, flip augment, constrained $M$ & 38.7 & 7.0\\
\hline
\end{tabular}
\end{center}
\caption{\textbf{Evaluation metrics by model.} The $+$MSE, $T_{new}$, coarse-to-fine, flip augment model yields the best PDJ and $L_2$.}
\label{table:metrics}
\end{table}

\begin{figure}[t]
    \centering
        \includegraphics[width=\linewidth]{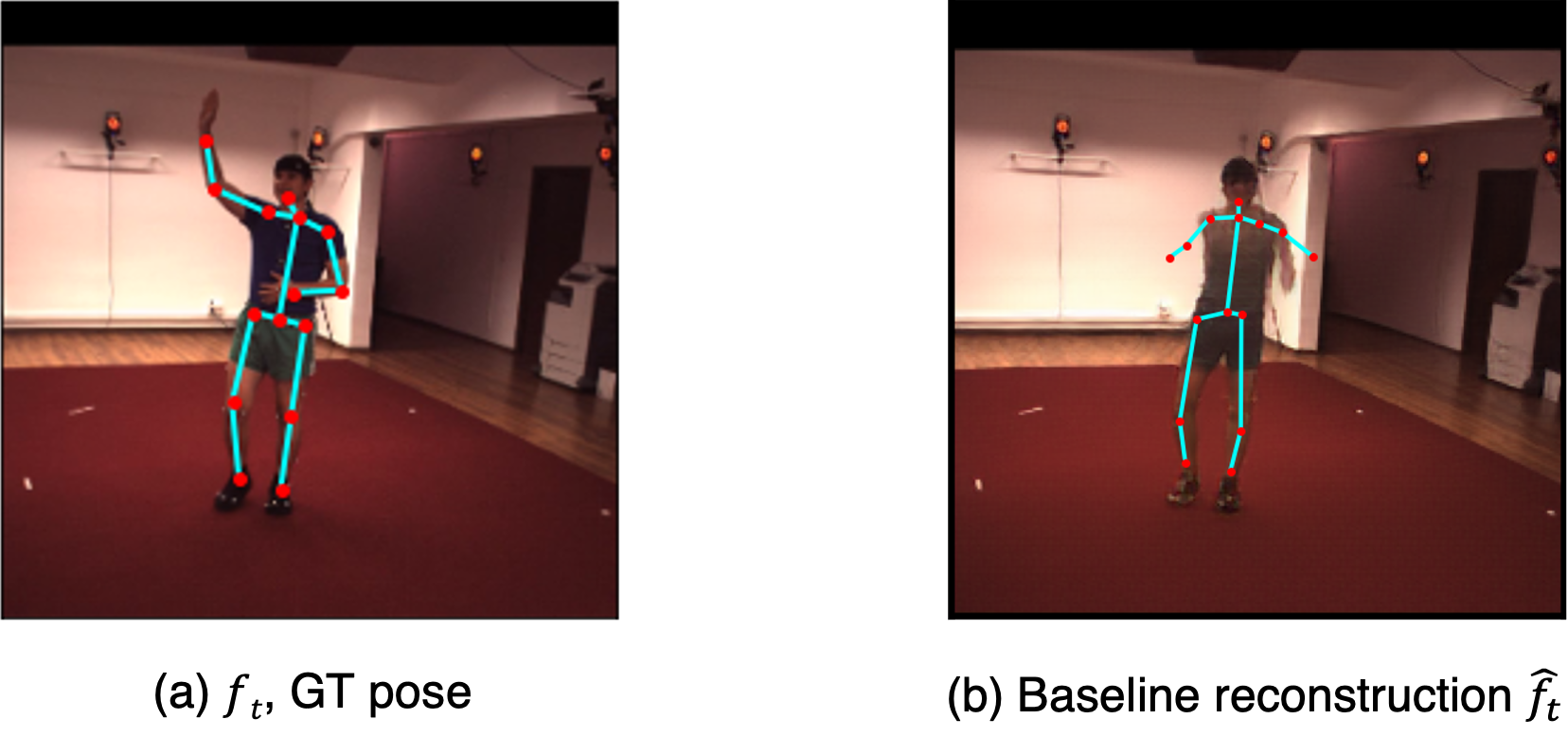}
\caption{\textbf{Reconstruction after 50 epochs.} The model is given only the image during training. \textbf{(a)} The ground truth pose shown for illustration purposes. \textbf{(b)} Baseline reconstruction of $f_t$.}
\label{fig:12}
\end{figure}

Next, we explore reconstruction quality in relation to pose estimate accuracy. Figure \ref{fig:12}b shows the baseline's reconstruction of $f_t$ after 50 epochs. The torso and legs show higher-quality reconstruction and keypoint grounding. The opposite is true for the elbow and wrists, which show lower-quality reconstruction and keypoint grounding. To improve overall keypoint grounding, we propose modifying the reconstruction loss by adding a pixel-wise loss term: MSE.

\begin{figure}[h]
    \centering
        \includegraphics[width=\linewidth]{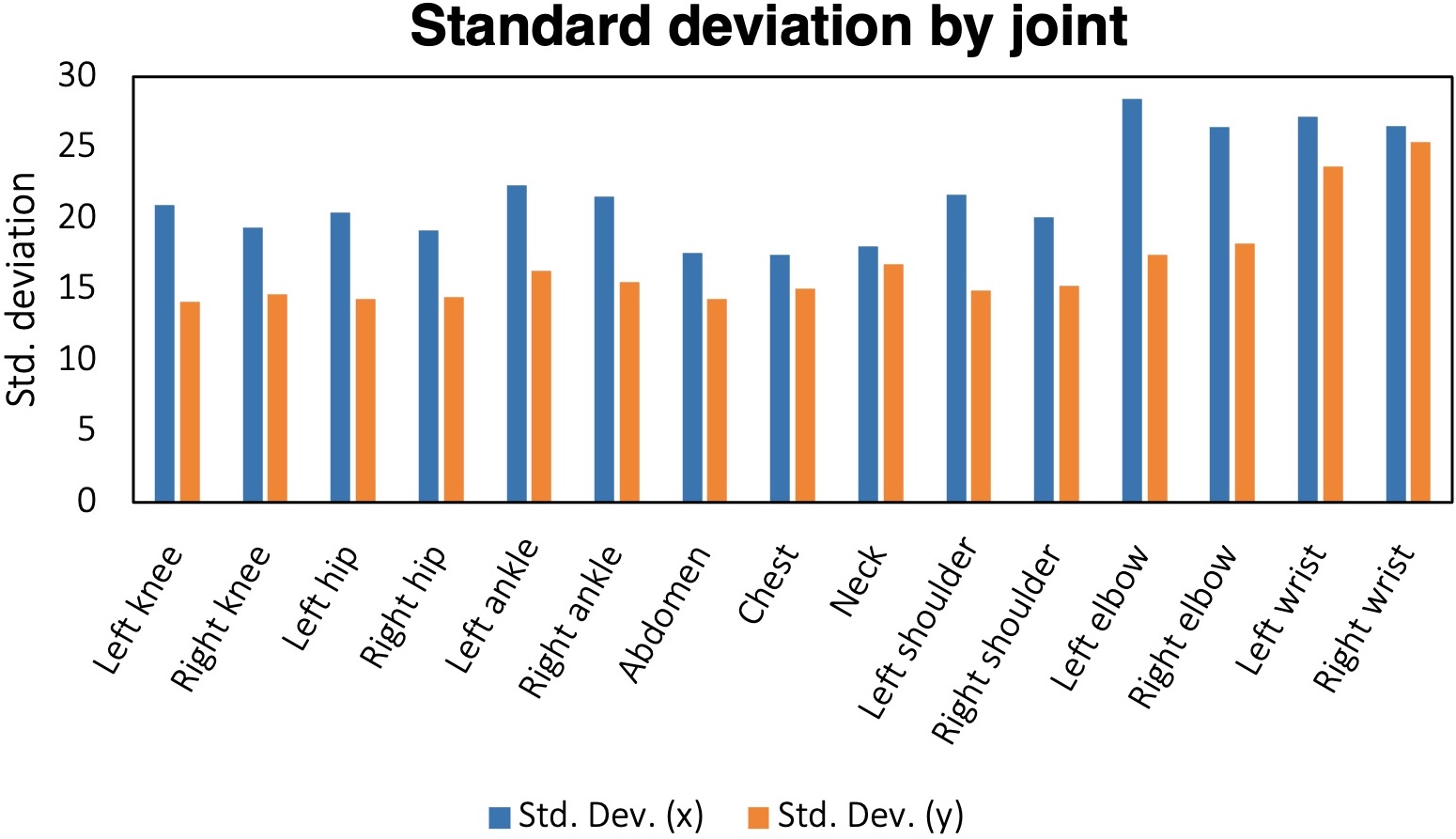}
\caption{\textbf{Standard deviation of $\boldsymbol{(x, y)}$ coordinates of ground truth joints.} The left and right elbow and wrist exhibit the most variation.}
\label{fig:limb_variation}
\end{figure}

One possible explanation for the disparity in reconstruction quality between the arms and legs comes from the distribution of the dataset---the left/right elbow and wrist exhibit the most variation in spatial configuration (Figure \ref{fig:limb_variation}).

\subsection{Results overall}

We now briefly compare our full model with \cite{schmidtke_unsupervised_2021}’s published checkpoint before diving into a detailed ablation study of the proposed changes. \cite{schmidtke_unsupervised_2021} attained a PDJ of 40.8 and $L_2$ error of 11.0 using 600K images trained over 30 epochs. In comparison, our best model ($+$MSE, $T_{new}$, coarse-to-fine, flip augment) reaches a PDJ of 42.6 and $L_2$ error of 6.4 using only 180K input examples trained over 50 epochs.

Furthermore, by the time our model reaches epoch 10, our $L_2$ is less than the checkpoint’s at 7.3. By epoch 30, our PDJ is comparable with the checkpoint’s at 40.4 (using three times fewer training samples) and by epoch 40, surpasses it at 41.8.

Analyzing Figure \ref{fig:1}, we observe that (1) our model predictions follow the contour of subject poses cleaner, seeing how the right wrist lies outside the subject in the fifth frame in Figure \ref{fig:1}a (\cite{schmidtke_unsupervised_2021}'s prediction), (2) our predicted keypoints for the neck and shoulder are less sunken, as observed in the third and fourth frame compared across Figures \ref{fig:1}a and \ref{fig:1}b, (3) our knee joint estimates are higher up (\ie generally more accurate), as observed in the first four frames compared across Figures \ref{fig:1}a and \ref{fig:1}b.

\subsection{Effect of the new reconstruction loss}

We now dive into the individual proposed changes and analyze them one-by-one. We begin by experimenting with the new reconstruction loss described in Section~\ref{sec:reconstructionloss}. Despite the initial hypothesis that improving reconstruction quality would improve pose estimates, we find this is not necessarily the case. Adding an MSE loss term speeds up reconstruction; but the model is worse at grounding keypoints, as shown by the lower reconstruction error accompanied by higher pose estimate error (Table \ref{table:recon_poseestimate}). In fact, the $+$MSE model exhibits worse performance than the baseline (PDJ$\Delta$:$-11.9$ points; $L_2\Delta$:$+2.0$) (Table \ref{table:metrics}). 

\begin{table}[t]
\begin{center}
\begin{tabular}{|c|c|c|}
\hline
\textbf{Model} & $\textbf{Error}_{\textbf{recon}}$ ($\boldsymbol{L_2}$) & $\textbf{Error}_\textbf{pose}$ ($\boldsymbol{L_2}$)\\
\hline
Baseline & 5414.5 & 7.2\\
\hline
$+$MSE & 4242.0 & 9.2\\
\hline
$+$MSE, $T_{new}$ & \textbf{3797.9} & \textbf{6.7}\\
\hline
\end{tabular}
\end{center}
\caption{\textbf{Comparison between reconstruction and pose estimate error after 50 epochs.} \textbf{(a) Baseline.} \textbf{(b) $\boldsymbol{+}$MSE, $\boldsymbol{T_{orig}}$.} Reconstruction error$\downarrow$, Pose estimate error$\uparrow$. \textbf{(c) $\boldsymbol{+}$MSE, $\boldsymbol{T_{new}}$.} Reconstruction error$\downarrow$, Pose estimate error$\downarrow$.}
\label{table:recon_poseestimate}
\end{table}

\begin{figure}[t]
    \centering
        \includegraphics[width=\linewidth]{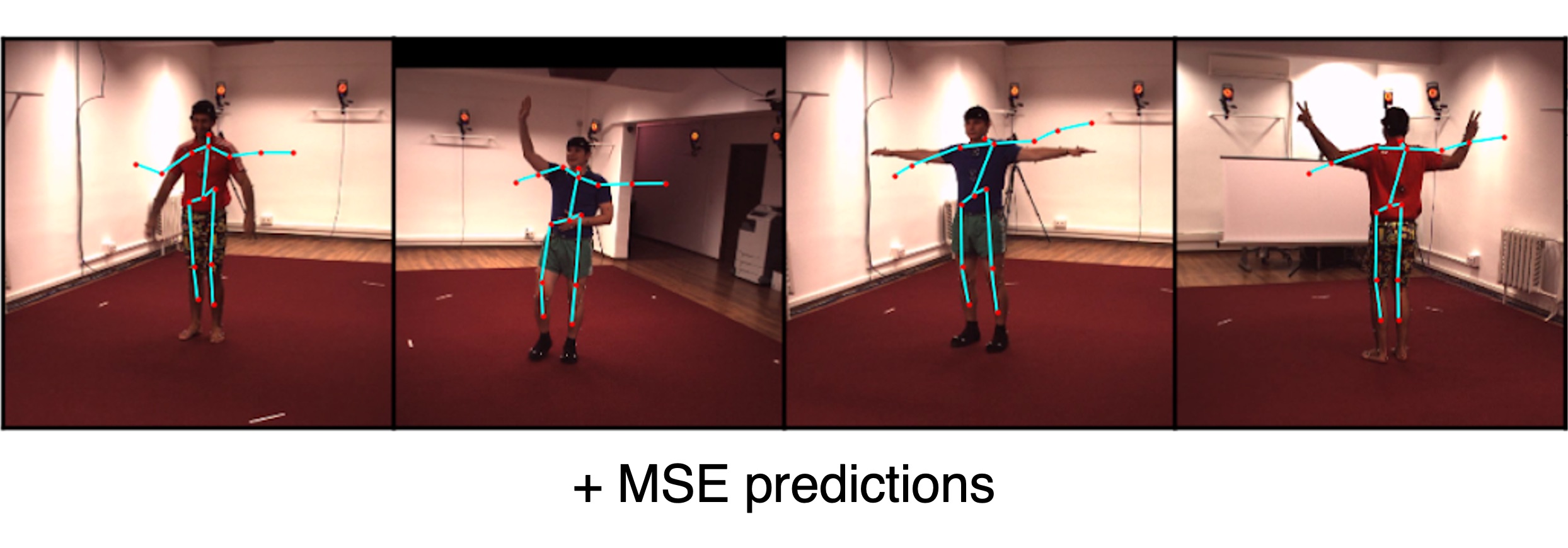}
\caption{\textbf{$\boldsymbol{+}$MSE model sample predictions.} The model consistently predicts an arms-out pose, regardless of the input frame $f_t$.}
\label{fig:31}
\end{figure}

Across the four sample predictions in Figure \ref{fig:31}, the leg joints are learned reasonably; however, the arms are consistently predicted to be extended outward, resembling the arms-out pose in template $T_{orig}$, regardless of whether the input frame contains a subject with arms up or down.

Notably, $T_{orig}$ (Figure \ref{fig:16}a) is highly unreflective of the pose distribution in the dataset. $T_{orig}$ represents an arms-out pose; however, the test dataset has nearly a $40:1$ ratio of arms-down poses to arms-extended (``arms-out'') poses. For simplicity, we assume the test and training set have a similar distribution, since the data was randomly sorted into training/test sets.

Our model is framed as a problem of template-matching (\ie distribution-matching) between estimate $T^*$ and reconstructed frame $\hat{f_t}$. In this context, adding an MSE term to the $l_{recon}$ encourages the model to reconstruct faster, and in the process, to quickly pick up correlations between the transformed template $T^*$ and original frame $f_t$. However, because the base template is so far from the data distribution, the model learns spurious correlations between $T^*$ and $\hat{f_t}$, resulting in misalignment between reconstruction and pose estimation.

In pushing the model toward faster, higher-quality reconstruction, it becomes even more important to provide an \textit{appropriate} template that reflects the underlying structure in the data, since the template provides a key inductive prior for learning meaningful relationships between pose estimation and reconstruction.

\subsection{Effect of the new template}

\begin{figure}[t]
    \centering
        \includegraphics[width=\linewidth]{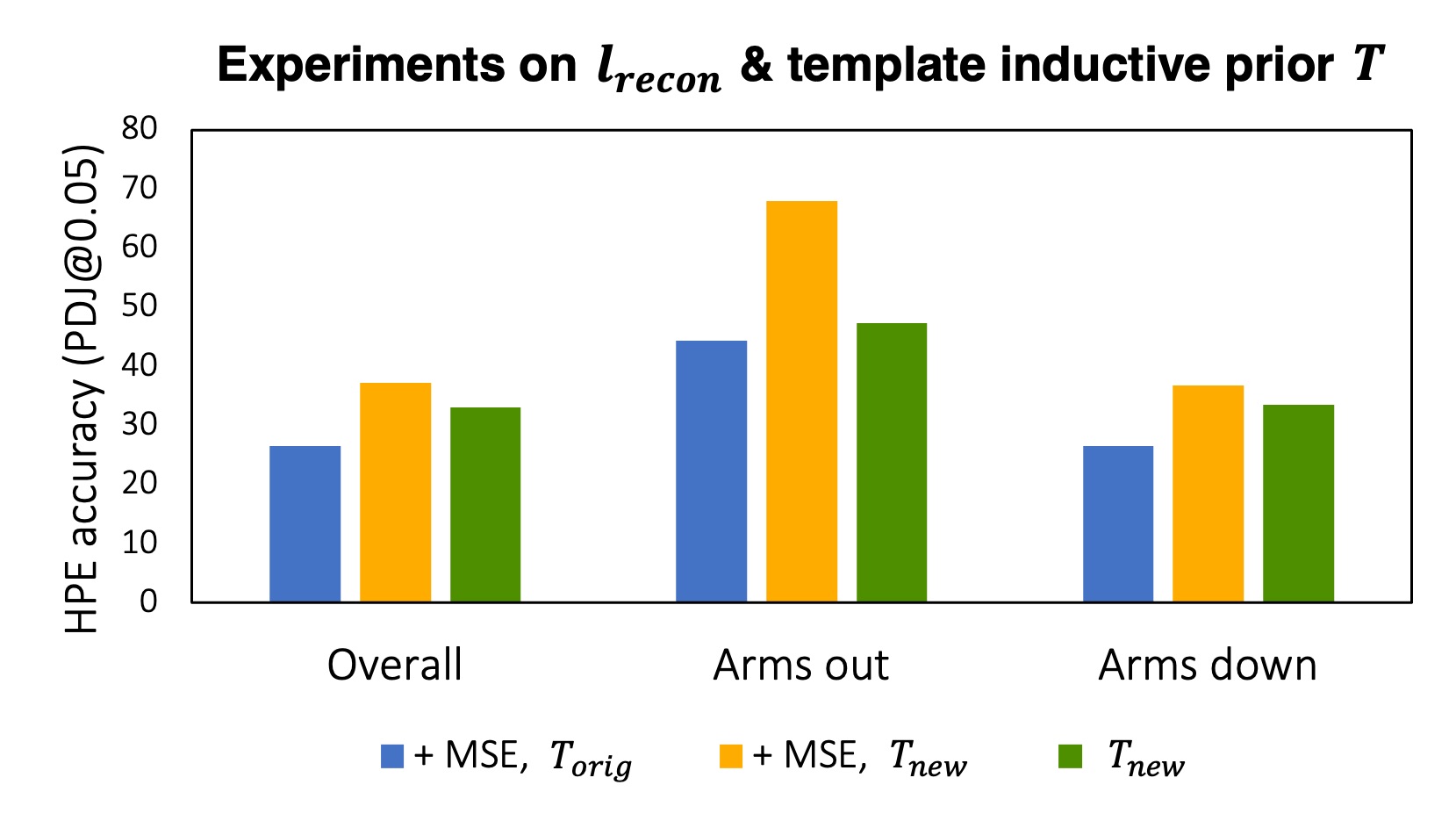}
\caption{\textbf{Effect of adopting $\boldsymbol{T_{new}}$ on pose estimate accuracy.} Adopting $T_{new}$ shows performance improvements when paired with the addition of an MSE loss term.}
\label{fig:30}
\end{figure}

Adopting the new template $T_{new}$ (Figure~\ref{fig:16}b) with MSE exhibits a strong lead over $T_{orig}$ with MSE (PDJ$\Delta$:$+10.6$ points; $L_2\Delta$:$-2.5$) and performs comparably with the baseline on PDJ and even outperforms the baseline's $L_2$ error. $T_{new}$ with MSE also outperforms using $T_{new}$ alone (PDJ$\Delta$:$+4.1$ points; $L_2\Delta$:$-0.8$). When $T_{new}$ is not paired with MSE (\ie the incentive to reconstruct faster), model performance drops overall, but most severely on arms-out poses (relative to $T_{new}$ paired with MSE) (Figure \ref{fig:30}).

Furthermore, in Table \ref{table:recon_poseestimate}, we see that the combination of MSE with $T_{new}$ reduces both reconstruction and pose estimate error (reconstruction error $\Delta$:$-1616.6$ points; pose estimation error $\Delta$:$-0.5$). Engineering an appropriate inductive prior is key to coordinating reconstruction with pose learning. Given the advantage of the MSE, $T_{new}$ combination, we use it as the new local benchmark against which we compare subsequent models. Next, we discuss our experiments updating the model to facilitate coarse-to-fine learning to refine the arm and forearm estimates.

\begin{figure*}[t]
\centering
\includegraphics[width=0.7\linewidth]{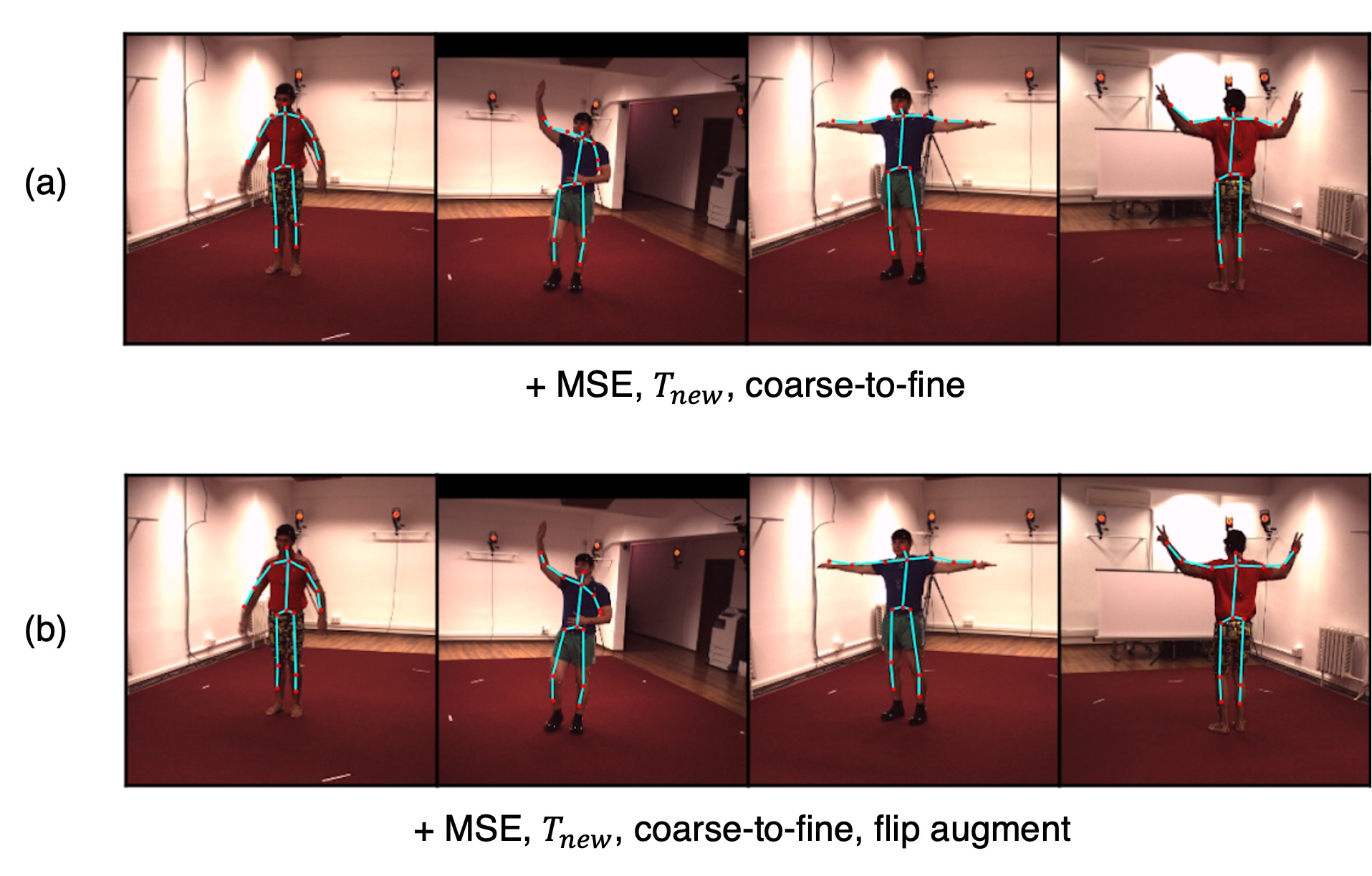}
\caption{\textbf{Sample predictions from models using coarse-to-fine learning, augmentation.} \textbf{(a)} $+$MSE, $T_{new}$, coarse-to-fine model. \textbf{(b)} $+$MSE, $T_{new}$, coarse-to-fine, flip augment model.}
\label{fig:27}
\end{figure*}

\begin{figure}[t]
    \centering
        \includegraphics[width=\linewidth]{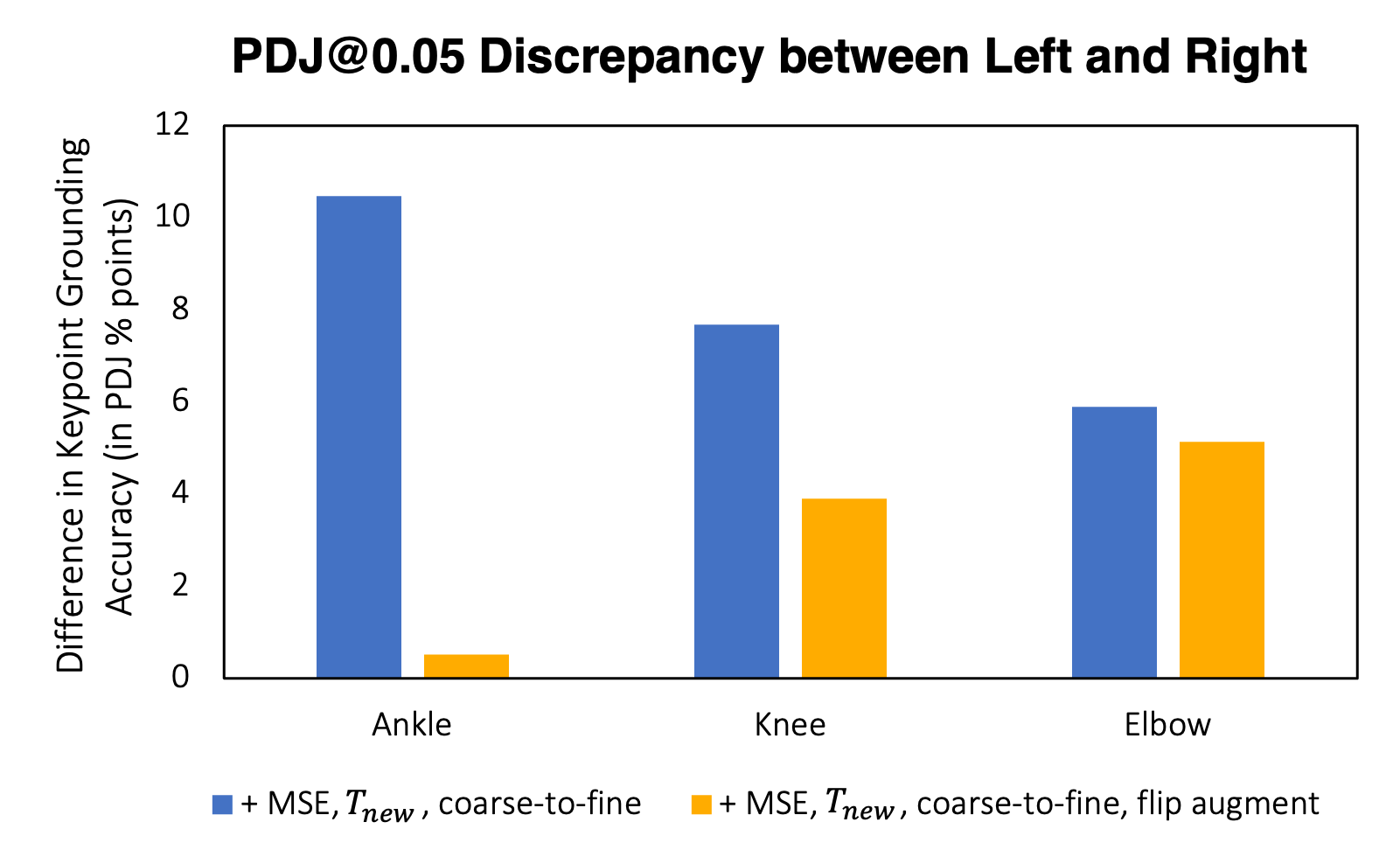}
\caption{\textbf{Difference in PDJ@0.05 between left and right-side ankle, knee, and elbow (the most extreme cases).} The discrepancy decreases after augmentation.}
\label{fig:23}
\end{figure}

\begin{figure}[t]
    \centering
        \includegraphics[width=\linewidth]{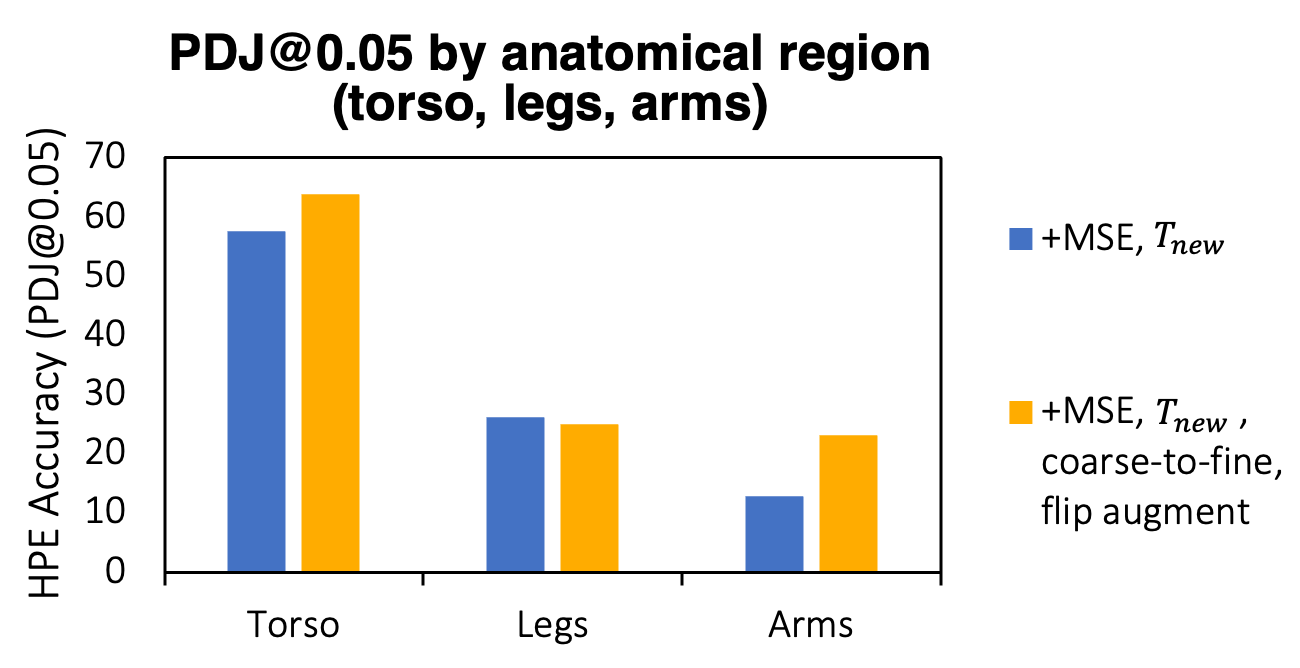}
\caption{\textbf{Model accuracy per anatomical region before/after coarse-to-fine learning and augmentation.} Torso keypoints: abdomen, chest, neck, hips, shoulders. Leg keypoints: knees, ankles. Arm keypoints: elbows, wrists.}
\label{fig:32}
\end{figure}

\subsection{Effect of coarse-to-fine learning}

Expanding $M$ and adapting a coarse-to-fine learning strategy described in Section~\ref{sec:coarsetofinelearning} yields PDJ$\Delta$:$+1.8$ points and $L_2\Delta$:$+0.3$ (Table \ref{table:metrics}). Examining the $L_2$ distribution, it seems that the slight increase in $L_2$ error comes from the increase in outliers (Figure \ref{fig:8}). On qualitative examples (Figure \ref{fig:27}a), coarse-to-fine learning yields pose estimates that closely follow subject contour.

One thing we notice in examining the per-joint accuracy deeper is the discrepancy in keypoint accuracy between left and right-side joints. For the ankle, knee, and elbow (the most extreme cases), the absolute differences are 10.5, 7.7, and 5.9 percentage points (Figure \ref{fig:23}). To account for natural differences in distribution between the two sides, we propose data diversification via augmentation.

\subsection{Effect of dataset augmentation}

We additionally augment the dataset by horizontal flipping as mentioned in Section~\ref{sec:datasetaugmentation}. To keep the training size to around 180K pairs, we take approximately half of the frames in the original training set and flip them, yielding a total of 181,728 training pairs $(f_t,f_{t+k})$, compared to 181,383, pre-augmentation.

After augmenting the dataset, the absolute differences in PDJ for the ankle, knee, and elbow drop to 0.5, 3.9, 5.1 percentage points, respectively (Figure \ref{fig:23}). The accuracy on arm keypoints jumps 10.2 percentage points (Figure~\ref{fig:32}). Overall PDJ jumps to 42.6 (the highest yet) and $L_2$ error drops to 6.4 (the lowest yet). To analyze whether this jump in performance is solely due to the addition of flip augmentation or whether it is a result of the specific combination of flip augmentation with coarse-to-fine learning, we also train a model using MSE, $T_{new}$, and flip augmentation with the original, unexpanded $M_{1 \leq i \leq 18}$. Compared with the $+$MSE, $T_{new}$ model, the performance drops when only augmentation is added (PDJ$\Delta$:$-2.9$ points; $L_2\Delta$:$+0.2$) (Table \ref{table:metrics}). Hence, we see that flip augmentation is working together with the expanded $M$ to produce more accurate predictions. 

With respect to outliers, we can see from the $L_2$ distribution (Figure \ref{fig:8}) that the combination of MSE, $T_{new}$, coarse-to-fine learning, and augmentation yields a model with one of the fewest number of outliers.

Despite its success, we notice inconsistent body part length proportions in some of its predictions. Similar to the baseline, the second and third frame in Figure \ref{fig:27}b depict significantly different predicted forearm lengths, despite showing the same subject. Next, we offer a new metric, \texttt{BPLP-C}, to measure this type of inconsistency and experiment with constraining $M$ to improve on this metric.

\subsection{Effect of constraining $\boldsymbol{M}$ for consistency}

\begin{figure}[t]
    \centering
            \includegraphics[width=\linewidth]{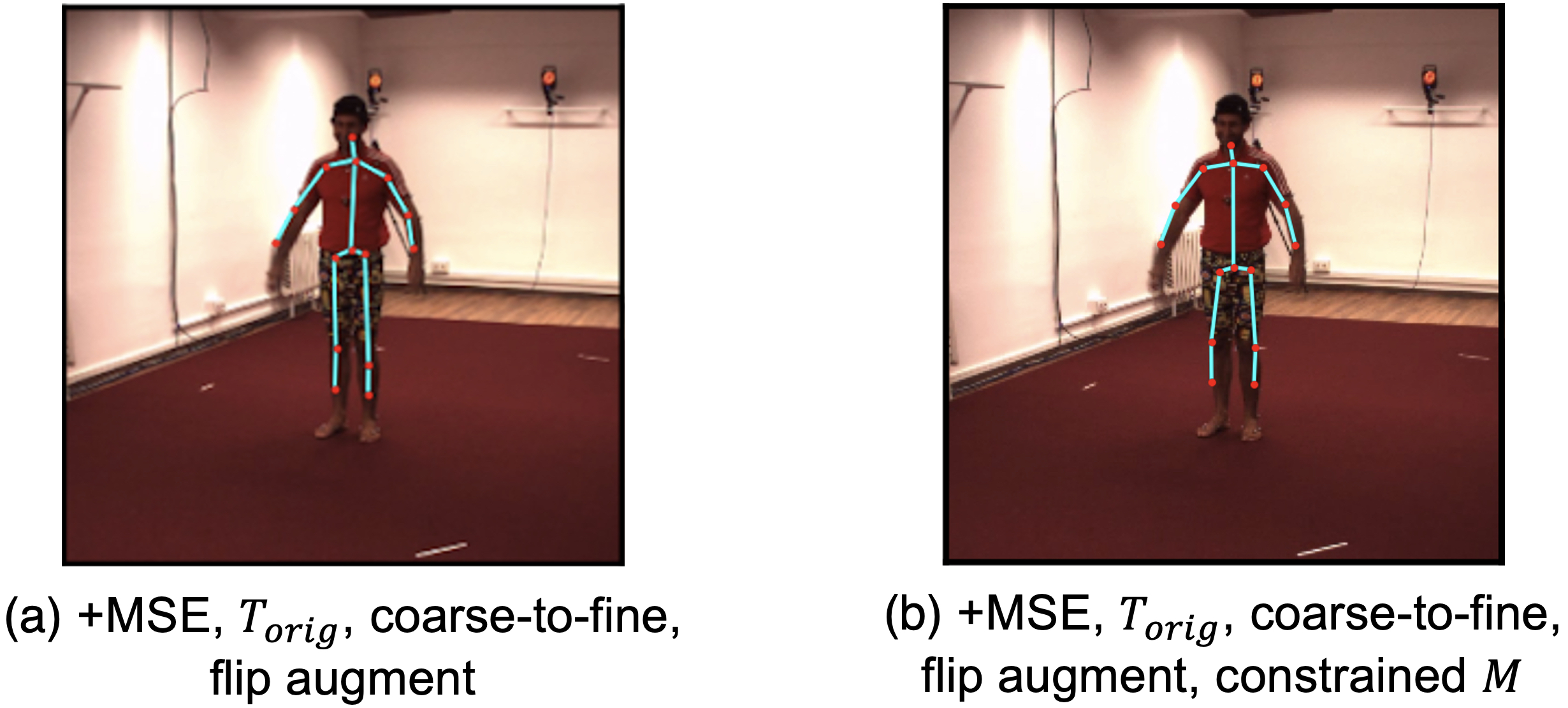}
\caption{\textbf{Sample prediction with/without constrained $\boldsymbol{M}$.} \textbf{(a)} Without constrained $M$. \textbf{(b)} With a constrained $M$, left limbs (e.g. left thigh) are relatively more consistent in length with right limbs (e.g. right thigh).}
\label{fig:33}
\end{figure}

Finally, we constrain $M$ for consistency as described in Section~\ref{sec:newconsistency}. Constraining $M$ yields a PDJ of 38.7 and $L_2$ error of 7.0, which is comparable with the MSE, $T_{new}$, coarse-to-fine model. Table \ref{table:bplp} shows \texttt{BPLP-C} by model. We notice that constraining $M$ yields the highest \texttt{BPLP-C} and more consistency between left and right limb lengths (Figure \ref{fig:33}b). This serves as a preliminary proof-of-concept that constraining the transformation matrix $M$ along these axes is a viable method to encourage consistent body part length proportions.

\begin{table}[t]
\begin{center}
\begin{tabular}{|m{6cm}|c|}
\hline
\multicolumn{1}{|c|}{\textbf{Model}} & \textbf{\texttt{BPLP-C}}\\
\hline
Published checkpoint & 2.03\\
\hline
Baseline & 9.82\\
\hline
$+$MSE, $T_{new}$, coarse-to-fine, flip augment & 5.17\\
\hline
$+$MSE, $T_{new}$, coarse-to-fine, flip augment, constrained $M$ & 12.32\\
\hline
\end{tabular}
\end{center}
\caption{\textbf{\texttt{BPLP} consistency across four models.} Constraining $M$ yields the highest \texttt{BPLP-C}.}
\label{table:bplp}
\end{table}

\section{Conclusion} 

In this paper, we present a modified self-supervised model pipeline that sets a new benchmark in PDJ@0.05 and normalized $L_2$ error for 2D human pose estimation that is based on a transformable, Gaussian shape template. Building off \cite{schmidtke_unsupervised_2021}, we make several new contributions:
\begin{enumerate}
    \item We analyze the settings in which reconstruction speed-up helps or hurts pose estimation and identify the importance of tuning the inductive prior to reflect some aspect of the data distribution, thereby enabling template-matching and coordinating reconstruction with pose estimation.
    \item By proposing ways to combine reconstruction loss, data augmentation, inductive prior tuning, and network-level adjustments, we find a model pipeline that exceeds baseline performance  using approximately three times less data.
    \item We propose \texttt{BPLP-C}, a metric that can be used to measure consistency in part length proportions in the absence of ground truth and propose one way to influence it---by constraining the transformation matrix.
\end{enumerate}

We consider a few ideas for extensions and future work. To improve our model performance further, it would be a good idea to incorporate some understanding of pose orientation into the model, so that it knows when the subject is facing away or toward the camera. It would be beneficial to test our pipeline on other HPE datasets to check for generalization. To relax the assumption we made about \texttt{BPLP}s being consistent across all subjects, we could create an end-to-end framework that learns not only the poses, but also, subject-specific templates based on a few different poses of the same subject. Furthermore, to measure the value of the \texttt{BPLP-C} metric, we hope to conduct human studies to understand how humans perceive predictions from models with higher \texttt{BPLP-C} in the 2D and 3D setting.

\section{Acknowledgements}
This work was done as part of NY's undergraduate senior thesis. We are grateful to Princeton Research Computing for providing the compute resources, and to the Princeton SEAS Howard B. Wentz, Jr. Junior Faculty Award (to OR) for enabling the publication and in-person presentation of this research.

{\small
\bibliographystyle{ieee_fullname}
\bibliography{egbib}

\begin{thebibliography}{10}\itemsep=-1pt

\bibitem{andriluka_2d_2014}
Mykhaylo Andriluka, Leonid Pishchulin, Peter Gehler, and Bernt Schiele.
\newblock 2d human pose estimation: New benchmark and state of the art analysis.
\newblock In {\em Proceedings of the IEEE Conference on Computer Vision and Pattern Recognition (CVPR)}, pages 3686--3693, 2014.

\bibitem{bauer_2023_weakly}
Peter Bauer, Arij Bouazizi, Ulrich Kressel, and Fabian~B. Flohr.
\newblock Weakly supervised multi-modal 3d human body pose estimation for autonomous driving.
\newblock In {\em 2023 IEEE Intelligent Vehicles Symposium (IV)}, pages 1--7, 2023.

\bibitem{berscheid2020self}
Lars Berscheid, Pascal Mei{\ss}ner, and Torsten Kr{\"o}ger.
\newblock Self-supervised learning for precise pick-and-place without object model.
\newblock {\em IEEE Robotics and Automation Letters}, 5(3):4828--4835, 2020.

\bibitem{bouazizi2021self}
Arij Bouazizi, Julian Wiederer, Ulrich Kressel, and Vasileios Belagiannis.
\newblock Self-supervised 3d human pose estimation with multiple-view geometry.
\newblock In {\em 2021 16th IEEE International Conference on Automatic Face and Gesture Recognition (FG 2021)}, pages 1--8, 2021.

\bibitem{cao_-bed_2022}
Ting Cao, Mohammad~Ali Armin, Simon Denman, Lars Petersson, and David Ahmedt-Aristizabal.
\newblock In-{Bed} {Human} {Pose} {Estimation} from {Unseen} and {Privacy}-{Preserving} {Image} {Domains}.
\newblock In {\em 2022 {IEEE} 19th {International} {Symposium} on {Biomedical} {Imaging} ({ISBI})}, pages 1--5, 2022.

\bibitem{deng2020self}
Xinke Deng, Yu Xiang, Arsalan Mousavian, Clemens Eppner, Timothy Bretl, and Dieter Fox.
\newblock Self-supervised 6d object pose estimation for robot manipulation.
\newblock In {\em 2020 IEEE International Conference on Robotics and Automation (ICRA)}, pages 3665--3671, 2020.

\bibitem{einfalt_uplift_2023}
Moritz Einfalt, Katja Ludwig, and Rainer Lienhart.
\newblock Uplift and upsample: Efficient 3d human pose estimation with uplifting transformers.
\newblock In {\em Proceedings of the IEEE/CVF Winter Conference on Applications of Computer Vision}, pages 2903--2913, 2023.

\bibitem{garau_capsulepose_2023}
Nicola Garau and Nicola Conci.
\newblock {CapsulePose}: {A} variational {CapsNet} for real-time end-to-end {3D} human pose estimation.
\newblock {\em Neurocomputing}, 523:81--91, 2023.

\bibitem{geng_densepose_2022}
Jiaqi Geng, Dong Huang, and Fernando De~la Torre.
\newblock Densepose from wifi.
\newblock {\em arXiv preprint arXiv:2301.00250}, 2022.

\bibitem{geng_human_2023}
Zigang Geng, Chunyu Wang, Yixuan Wei, Ze Liu, Houqiang Li, and Han Hu.
\newblock Human pose as compositional tokens.
\newblock In {\em Proceedings of the IEEE Conference on Computer Vision and Pattern Recognition (CVPR)}, pages 660--671, 2023.

\bibitem{ghodrati_mr_2019}
Vahid Ghodrati, Jiaxin Shao, Mark Bydder, Ziwu Zhou, Wotao Yin, Kim-Lien Nguyen, Yingli Yang, and Peng Hu.
\newblock {MR} image reconstruction using deep learning: evaluation of network structure and loss functions.
\newblock {\em Quantitative Imaging in Medicine and Surgery}, 9(9):1516--1527, 2019.

\bibitem{gholami2022self}
Mohsen Gholami, Ahmad Rezaei, Helge Rhodin, Rabab Ward, and Z~Jane Wang.
\newblock Self-supervised 3d human pose estimation from video.
\newblock {\em Neurocomputing}, 488:97--106, 2022.

\bibitem{ionescu_human36m_2014}
Catalin Ionescu, Dragos Papava, Vlad Olaru, and Cristian Sminchisescu.
\newblock Human3.{6M}: {Large} {Scale} {Datasets} and {Predictive} {Methods} for {3D} {Human} {Sensing} in {Natural} {Environments}.
\newblock {\em IEEE Transactions on Pattern Analysis and Machine Intelligence}, 36(7):1325--1339, 2014.

\bibitem{johnson_perceptual_2016}
Justin Johnson, Alexandre Alahi, and Li Fei-Fei.
\newblock Perceptual {Losses} for {Real}-{Time} {Style} {Transfer} and {Super}-{Resolution}.
\newblock In {\em Computer {Vision} – {ECCV} 2016}, pages 694--711, 2016.

\bibitem{Kocabas_2019_self}
Muhammed Kocabas, Salih Karagoz, and Emre Akbas.
\newblock Self-supervised learning of 3d human pose using multi-view geometry.
\newblock In {\em Proceedings of the IEEE Conference on Computer Vision and Pattern Recognition (CVPR)}, pages 1077--1086, 2019.

\bibitem{kress_2019_human}
Viktor Kress, Janis Jung, Stefan Zernetsch, Konrad Doll, and Bernhard Sick.
\newblock Human pose estimation in real traffic scenes.
\newblock In {\em 2018 IEEE Symposium Series on Computational Intelligence (SSCI)}, pages 518--523, 2018.

\bibitem{Kundu_2020_self}
Jogendra~Nath Kundu, Siddharth Seth, Varun Jampani, Mugalodi Rakesh, R~Venkatesh Babu, and Anirban Chakraborty.
\newblock Self-supervised 3d human pose estimation via part guided novel image synthesis.
\newblock In {\em Proceedings of the IEEE Conference on Computer Vision and Pattern Recognition (CVPR)}, pages 6152--6162, 2020.

\bibitem{Kundu_2022_uncertainty}
Jogendra~Nath Kundu, Siddharth Seth, Pradyumna YM, Varun Jampani, Anirban Chakraborty, and R.~Venkatesh Babu.
\newblock Uncertainty-aware adaptation for self-supervised 3d human pose estimation.
\newblock In {\em Proceedings of the IEEE/CVF Conference on Computer Vision and Pattern Recognition (CVPR)}, pages 20448--20459, 2022.

\bibitem{ledig_photo-realistic_2017}
Christian Ledig, Lucas Theis, Ferenc Huszar, Jose Caballero, Andrew Cunningham, Alejandro Acosta, Andrew Aitken, Alykhan Tejani, Johannes Totz, Zehan Wang, and Wenzhe Shi.
\newblock Photo-realistic single image super-resolution using a generative adversarial network.
\newblock In {\em Proceedings of the IEEE Conference on Computer Vision and Pattern Recognition (CVPR)}, pages 4681--4690, 2017.

\bibitem{lee_hupr_2023}
Shih-Po Lee, Niraj~Prakash Kini, Wen-Hsiao Peng, Ching-Wen Ma, and Jenq-Neng Hwang.
\newblock Hupr: A benchmark for human pose estimation using millimeter wave radar.
\newblock In {\em Proceedings of the IEEE/CVF Winter Conference on Applications of Computer Vision}, pages 5715--5724, 2023.

\bibitem{li_geometry-driven_2020}
Yang Li, Kan Li, Shuai Jiang, Ziyue Zhang, Congzhentao Huang, and Richard Yi~Da Xu.
\newblock Geometry-{Driven} {Self}-{Supervised} {Method} for {3D} {Human} {Pose} {Estimation}.
\newblock {\em Proceedings of the AAAI Conference on Artificial Intelligence}, 34(07):11442--11449, 2020.

\bibitem{li_human_2022}
Yanping Li, Ruyi Liu, Xiangyang Wang, and Rui Wang.
\newblock Human pose estimation based on lightweight basicblock.
\newblock {\em Machine Vision and Applications}, 34(1):3, 2022.

\bibitem{lin2014microsoft}
Tsung-Yi Lin, Michael Maire, Serge Belongie, James Hays, Pietro Perona, Deva Ramanan, Piotr Doll{\'a}r, and C.~Lawrence Zitnick.
\newblock Microsoft coco: Common objects in context.
\newblock In {\em Computer Vision -- ECCV 2014}, pages 740--755, 2014.

\bibitem{ma2021context}
Xiaoxuan Ma, Jiajun Su, Chunyu Wang, Hai Ci, and Yizhou Wang.
\newblock Context modeling in 3d human pose estimation: A unified perspective.
\newblock In {\em Proceedings of the IEEE Conference on Computer Vision and Pattern Recognition (CVPR)}, pages 6238--6247, 2021.

\bibitem{osokin_real-time_2018}
Daniil Osokin.
\newblock Real-time 2d multi-person pose estimation on cpu: Lightweight openpose.
\newblock {\em arXiv preprint arXiv:1811.12004}, 2018.

\bibitem{qian_hstformer_2023}
Xiaoye Qian, Youbao Tang, Ning Zhang, Mei Han, Jing Xiao, Ming-Chun Huang, and Ruei-Sung Lin.
\newblock Hstformer: Hierarchical spatial-temporal transformers for 3d human pose estimation.
\newblock {\em arXiv preprint arXiv:2301.07322}, 2023.

\bibitem{richardson_learning_2017}
Elad Richardson, Matan Sela, Roy Or-El, and Ron Kimmel.
\newblock Learning {Detailed} {Face} {Reconstruction} from a {Single} {Image}.
\newblock In {\em Proceedings of the IEEE Conference on Computer Vision and Pattern Recognition (CVPR)}, pages 5553--5562, 2017.

\bibitem{santavas_attention_2021}
Nicholas Santavas, Ioannis Kansizoglou, Loukas Bampis, Evangelos Karakasis, and Antonios Gasteratos.
\newblock Attention! {A} {Lightweight} {2D} {Hand} {Pose} {Estimation} {Approach}.
\newblock {\em IEEE Sensors Journal}, 21(10):11488--11496, 2021.

\bibitem{schmidtke_unsupervised_2021}
Luca Schmidtke, Athanasios Vlontzos, Simon Ellershaw, Anna Lukens, Tomoki Arichi, and Bernhard Kainz.
\newblock Unsupervised human pose estimation through transforming shape templates.
\newblock In {\em Proceedings of the IEEE Conference on Computer Vision and Pattern Recognition (CVPR)}, pages 2484--2494, 2021.

\bibitem{sharma2022neural}
Prafull Sharma, Ayush Tewari, Yilun Du, Sergey Zakharov, Rares~Andrei Ambrus, Adrien Gaidon, William~T Freeman, Fredo Durand, Joshua~B Tenenbaum, and Vincent Sitzmann.
\newblock Neural groundplans: Persistent neural scene representations from a single image.
\newblock In {\em The Eleventh International Conference on Learning Representations}, 2022.

\bibitem{shu_feature-metric_2020}
Chang Shu, Kun Yu, Zhixiang Duan, and Kuiyuan Yang.
\newblock Feature-{Metric} {Loss} for {Self}-supervised {Learning} of {Depth} and {Egomotion}.
\newblock In {\em Computer {Vision} – {ECCV} 2020}, pages 572--588, 2020.

\bibitem{snell_learning_2017}
Jake Snell, Karl Ridgeway, Renjie Liao, Brett~D. Roads, Michael~C. Mozer, and Richard~S. Zemel.
\newblock Learning to generate images with perceptual similarity metrics.
\newblock In {\em 2017 {IEEE} {International} {Conference} on {Image} {Processing} ({ICIP})}, pages 4277--4281, 2017.

\bibitem{sosa2023self}
Jose Sosa and David Hogg.
\newblock Self-supervised 3d human pose estimation from a single image.
\newblock In {\em Proceedings of the IEEE Conference on Computer Vision and Pattern Recognition (CVPR)}, pages 4787--4796, 2023.

\bibitem{tendle_study_2021}
Atharva Tendle and Mohammad~Rashedul Hasan.
\newblock A study of the generalizability of self-supervised representations.
\newblock {\em Machine Learning with Applications}, 6:100124, 2021.

\bibitem{wan_self-supervised_2019}
Chengde Wan, Thomas Probst, Luc~Van Gool, and Angela Yao.
\newblock Self-supervised 3d hand pose estimation through training by fitting.
\newblock In {\em Proceedings of the IEEE/CVF Conference on Computer Vision and Pattern Recognition (CVPR)}, pages 10853--10862, 2019.

\bibitem{Wandt_2021_CVPR}
Bastian Wandt, Marco Rudolph, Petrissa Zell, Helge Rhodin, and Bodo Rosenhahn.
\newblock Canonpose: Self-supervised monocular 3d human pose estimation in the wild.
\newblock In {\em Proceedings of the IEEE/CVF Conference on Computer Vision and Pattern Recognition (CVPR)}, pages 13294--13304, 2021.

\bibitem{wang_leverage_2019}
Sijia Wang, Fabian~B. Flohr, Hui Xiong, Tuopu Wen, Baofeng Wang, Mengmeng Yang, and Diange Yang.
\newblock Leverage of {Limb} {Detection} in {Pose} {Estimation} for {Vulnerable} {Road} {Users}.
\newblock In {\em 2019 {IEEE} {Intelligent} {Transportation} {Systems} {Conference} ({ITSC})}, pages 528--534, 2019.

\bibitem{wang_multiscale_2003}
Z. Wang, E.P. Simoncelli, and A.C. Bovik.
\newblock Multiscale structural similarity for image quality assessment.
\newblock In {\em The {Thirty}-{Seventh} {Asilomar} {Conference} on {Signals}, {Systems} \& {Computers}, 2003}, volume~2, pages 1398--1402 Vol.2, 2003.

\bibitem{xu_can_2023}
Dingning Xu, Lijun Guo, Rong Zhang, Jiangbo Qian, and Shangce Gao.
\newblock Can relearning local representation help small networks for human pose estimation?
\newblock {\em Neurocomputing}, 518:418--430, 2023.

\bibitem{xu_vitpose_2022}
Yufei Xu, Jing Zhang, Qiming Zhang, and Dacheng Tao.
\newblock {ViTPose}: {Simple} {Vision} {Transformer} {Baselines} for {Human} {Pose} {Estimation}.
\newblock {\em Advances in Neural Information Processing Systems}, 35:38571--38584, 2022.

\bibitem{yang_camerapose_2023}
Cheng-Yen Yang, Jiajia Luo, Lu Xia, Yuyin Sun, Nan Qiao, Ke Zhang, Zhongyu Jiang, Jenq-Neng Hwang, and Cheng-Hao Kuo.
\newblock Camerapose: Weakly-supervised monocular 3d human pose estimation by leveraging in-the-wild 2d annotations.
\newblock In {\em Proceedings of the IEEE/CVF Winter Conference on Applications of Computer Vision}, pages 2924--2933, 2023.

\bibitem{zanfir_2023_hum3dil}
Andrei Zanfir, Mihai Zanfir, Alex Gorban, Jingwei Ji, Yin Zhou, Dragomir Anguelov, and Cristian Sminchisescu.
\newblock Hum3dil: Semi-supervised multi-modal 3d humanpose estimation for autonomous driving.
\newblock In {\em Proceedings of The 6th Conference on Robot Learning}, volume 205, pages 1114--1124, 2023.

\bibitem{zeng_df2net_2019}
Xiaoxing Zeng, Xiaojiang Peng, and Yu Qiao.
\newblock Df2net: A dense-fine-finer network for detailed 3d face reconstruction.
\newblock In {\em Proceedings of the IEEE/CVF International Conference on Computer Vision (ICCV)}, pages 2315--2324, 2019.

\bibitem{zhang_deblurring_2020}
Kaihao Zhang, Wenhan Luo, Yiran Zhong, Lin Ma, Bjorn Stenger, Wei Liu, and Hongdong Li.
\newblock Deblurring by realistic blurring.
\newblock In {\em Proceedings of the IEEE Conference on Computer Vision and Pattern Recognition (CVPR)}, pages 2737--2746, 2020.

\bibitem{zhang_neuromorphic_2023}
Zhongyang Zhang, Kaidong Chai, Haowen Yu, Ramzi Majaj, Francesca Walsh, Edward Wang, Upal Mahbub, Hava Siegelmann, Donghyun Kim, and Tauhidur Rahman.
\newblock Neuromorphic high-frequency 3d dancing pose estimation in dynamic environment.
\newblock {\em Neurocomputing}, page 126388, 2023.

\bibitem{zhang_simple_2020}
Zhe Zhang, Jie Tang, and Gangshan Wu.
\newblock Simple and lightweight human pose estimation.
\newblock {\em arXiv preprint arXiv:1911.10346}, 2020.

\bibitem{zhao_dpit_2022}
Shuaitao Zhao, Kun Liu, Yuhang Huang, Qian Bao, Dan Zeng, and Wu Liu.
\newblock {DPIT}: {Dual}-{Pipeline} {Integrated} {Transformer} for {Human} {Pose} {Estimation}.
\newblock In {\em Artificial {Intelligence}}, pages 559--576, 2022.

\bibitem{Zheng_2022_multimodal}
Jingxiao Zheng, Xinwei Shi, Alexander Gorban, Junhua Mao, Yang Song, Charles~R. Qi, Ting Liu, Visesh Chari, Andre Cornman, Yin Zhou, Congcong Li, and Dragomir Anguelov.
\newblock Multi-modal 3d human pose estimation with 2d weak supervision in autonomous driving.
\newblock In {\em Proceedings of the IEEE/CVF Conference on Computer Vision and Pattern Recognition (CVPR) Workshops}, pages 4478--4487, 2022.

\end{thebibliography}
}

\end{document}